\documentclass[letterpaper,10pt,conference]{ieeeconf}
\usepackage{cite}
\usepackage{graphicx}
\usepackage{subcaption}
\usepackage[cmex10]{amsmath}
\usepackage{multirow}
\usepackage{caption}
\usepackage{booktabs} 
\usepackage{algorithmicx}
\usepackage[section]{algorithm}
\usepackage{algpascal}
\usepackage{algc}
\usepackage{algcompatible}
\usepackage{algpseudocode}
\usepackage{xcolor}
\usepackage{array}
\usepackage{url}
\usepackage{bbm}
\def \*{\star}
\usepackage{todonotes}

\usepackage{epsfig}
\usepackage{amsfonts,amssymb,multirow,bigstrut,booktabs,ctable,latexsym}
\usepackage{mathtools}
\usepackage[mathscr]{eucal}
\usepackage{verbatim}

\usepackage{amsthm}
\usepackage{algorithm}
\usepackage[normalem]{ulem}
\newtheorem{definition}{Definition}

\newtheorem{theorem}{Theorem}

\newtheorem{remark}{Remark}
\newtheorem{assumption}{Assumption}

\DeclareMathAlphabet{\pazocal}{OMS}{zplm}{m}{n}
\newcommand{\A}{\pazocal{A}}

\newtheorem{prop}[theorem]{Proposition}

\IEEEoverridecommandlockouts

\title{A Method for Fast Autonomy Transfer in Reinforcement Learning}
	
	\author{Dinuka Sahabandu$^{1}$, Bhaskar Ramasubramanian$^{2}$, Michail Alexiou$^{3}$,\\ J. Sukarno Mertoguno$^{4}$, Linda Bushnell$^{1}$, Radha Poovendran$^{1}$%
		\thanks{$^{1}$Network Security Lab, Department of Electrical and Computer Engineering, 
			University of Washington, Seattle, WA 98195, USA. \newline
			{\tt\small \{sdinuka, lb2, rp3\}@uw.edu}}
\thanks{$^2$Electrical and Computer Engineering, Western Washington University, Bellingham, WA 98225, USA.
			{\tt ramasub@wwu.edu}}
			\thanks{$^3$Department of Computer Science, Kennesaw State University, Marietta, GA 30060, USA.
			{\tt malexiou@kennesaw.edu}}
			\thanks{$^4$ School of Cybersecurity and Privacy, Georgia Institute of Technology, Atlanta, GA 30332.
			{\tt karno@gatech.edu}}
	}
\begin{document}	
\maketitle

\begin{abstract}
This paper introduces 
a novel reinforcement learning (RL) strategy designed to facilitate rapid autonomy transfer by utilizing pre-trained critic value functions from multiple environments. Unlike traditional methods that require extensive retraining or fine-tuning, our approach integrates existing knowledge, enabling an RL agent to adapt swiftly to new settings without requiring extensive computational resources. Our contributions include development of the {Multi-critic Actor-Critic (MCAC)}  algorithm, establishing its convergence, and empirical evidence demonstrating its efficacy. Our experimental results show that MCAC significantly outperforms the baseline Actor-Critic (AC) algorithm, achieving up to $22.76 \times$ faster autonomy transfer and higher reward accumulation. This advancement underscores the potential of leveraging accumulated knowledge for efficient adaptation in RL applications.
\end{abstract}

\section{Introduction}\label{sec:Introduction}

In the realms of artificial intelligence and data-driven control, reinforcement learning (RL) stands out as a powerful paradigm for solving complex decision-making problems in unknown or unseen environments \cite{sutton2018reinforcement}. RL involves an agent learning to make decisions by interacting with an environment, by aiming to maximize a cumulative reward signal provided by the environment over time. This process of trial and error, coupled with the feedback received through rewards, allows the agent to develop a strategy or policy that guides its actions towards achieving its goals.

RL has been successfully applied in multiple domains, including robotics, autonomous vehicles, games, and mobile networks \cite{hafner2011reinforcement, mnih2015human, silver2016mastering, zhang2019deep, sadigh2016planning, yan2018data, you2019advanced}.
%
Despite these successes, RL faces a significant challenge when an agent encounters new environments. In such a scenario, the agent must learn from the very beginning in each new setting, a process that can be prohibitively time-consuming and resource-intensive. The ability to transfer knowledge—utilizing insights gained from previous experiences to expedite learning in new situations—has therefore become a key area of focus. Knowledge transfer can dramatically reduce the amount of interaction needed with a new environment to reach optimal or near-optimal performance, thereby accelerating the learning process and reducing computational costs. 


Yet, effectively achieving this transfer of knowledge poses challenges. In deep RL \cite{arulkumaran2017deep}, where agents use deep neural networks to approximate policies or value functions, techniques such as transfer learning and fine-tuning have shown promise \cite{zhu2023transfer, campos2021beyond}. While transfer learning and fine-tuning techniques offer some solutions, they typically involve re-training parts of the network. This process, albeit not as extensive as retraining the entire network, still demands considerable computational effort and time. 
On the other hand, traditional RL paradigms operating in discrete environments which use a tabular approach to storing value functions (e.g., Q-learning \cite{watkins1992q}, presents a different challenge- \emph{how can pre-learned knowledge be leveraged in discrete environments without complex neural network architectures used in deep RL?}. 


In this paper, we investigate the concept of rapid autonomy transfer within the scope of classical reinforcement learning. Specifically, we examine this problem through the lens of Actor-Critic (AC) algorithms under a discounted reward framework. 
AC methods combine policy-based decision making (actor) with value-based evaluation (critic), offering a balanced approach to learning by directly integrating feedback on the quality of actions taken \cite{konda1999actor}. 
Actor-Critic methods are distinguished by their rapid convergence properties and minimal convergence errors, closely approaching optimal performance. The discounted reward setting, central to our analysis, emphasizes the importance of future rewards by applying a discount factor, which ensures that immediate gains do not overshadow long-term benefits.

We propose a novel approach to enable rapid autonomy transfer in RL that uses pre-trained critic value functions from multiple environments and combines them to assist in training a new actor for a different environment. 
Our approach 
bypasses the need for retraining or partially training critics by focusing on the optimal integration of existing knowledge, effectively leveraging the knowledge accumulated from multiple critics to navigate new environments. 

Specifically, our \textbf{Multi-critic Actor-Critic (MCAC)} learns a set of weights to modulate influence of pre-trained value functions from diverse environments.
By leveraging pre-existing value functions without the need for retraining, MCAC not only conserves computational resources but also accelerates adaptation to new environments. This contribution marks a significant advancement in the application of RL techniques, opening up new possibilities for their use in diverse and dynamic settings. We make the following contributions in this paper:
\begin{itemize}
    \item We introduce the MCAC algorithm to enable fast autonomy transfer in RL. 
    \item We establish convergence of weights and policies learned by our MCAC algorithm.
    \item Experimentally, we show that MCAC achieves autonomy transfer up to \textbf{22.76x} times faster than a baseline AC algorithm, while also achieving higher rewards.
\end{itemize}

The rest of this paper is organized as: Sec. \ref{sec:RelatedWork} describes related work and Sec. \ref{sec:Preliminaries} presents necessary preliminaries. Sec. \ref{sec:ProblemSetup} introduces our MCAC approach and presents our main results. 
Sec. \ref{sec:Experiments} reports results of experimental evaluations of MCAC and Sec. \ref{sec:Conclusion} concludes the paper. 


\section{Related Work}\label{sec:RelatedWork}

This section briefly describes related work in multi-critic learning, and methods to aggregate information from critics.  

Multi-Critic Actor Learning was proposed in \cite{mysore2021multi} in the context of multi-task learning as an alternative to maintaining separate critics for each task being trained while training a single multi-task actor. Explicitly distinguishing between tasks also eliminates a need for critics to learn to do so and mitigates interference between task-value estimates. However, while \cite{mysore2021multi} seeks to `select' a particular style of task satisfaction in a given environment, our MCAC approach `combines' multiple styles of task satisfaction to execute a given task in a new environment. 
Recently, the authors of \cite{li2023multi} introduced a multi-actor mechanism that combined novel exploration strategies together with a Q-value weighting technique to accelerate learning of optimal behaviors. 
Different from \cite{li2023multi}, where experiments for any given environment provide comparisons against benchmarks within the same environment, our goal in this paper is to accomplish autonomy transfer across different environments. 

Reward machines \cite{icarte2022reward} provide a structured framework for defining, formalizing, and managing rewards in an intuitive and human-understandable manner. 
Structuring rewards in this way makes it possible to guide RL agents more effectively. 
Reward machines were used to inform design of a Q-learning algorithm in finite state-action environments in \cite{icarte2018using}, and extended to partially observable environments in \cite{icarte2023learning}. 
However, the current state-of-the-art in this domain does not provide a direct way to encode and aggregate information from disparate sources and do not have a mechanism to encode distracting rewards. 
In comparison, our MCAC technique aggregates information from multiple scenarios and uses this information to achieve significant speedup in terms of run time and number of learning episodes required. 

\section{Preliminaries}\label{sec:Preliminaries}

This section introduces necessary preliminaries on Markov decision processes (MDPs), reinforcement learning (RL), and stochastic approximation (SA). 

\subsection{MDPs and RL}

Let $(\Omega, \mathcal{F}, \mathcal{P})$ denote a probability space, where $\Omega$ is a sample space, $\mathcal{F}$ is a $\sigma-$algebra of subsets of $\Omega$, and $\mathcal{P}$ is a probability measure on $\mathcal{F}$. 
A random variable is a map $Y: \Omega \rightarrow \mathbb{R}$. 
We assume that the environment of the RL agent is described by a Markov decision process (MDP) \cite{puterman2014markov}. 

\begin{definition}
An MDP is a tuple $\mathcal{M}:= (S, A, P, r, \gamma)$, where $S$ is a finite set of states, $A$ is a finite set of actions, and 
$P(s'|s,a)$ is the probability of transiting to state $s'$ when action $a$ is taken in state $s$. 
$r: S \times A \rightarrow \mathbb{R}$ is the reward obtained by the agent when it takes action $a$ in state $s$. 
$\gamma \in (0,1]$ is a discounting factor.
\end{definition}

An RL agent typically does not have knowledge of 
$P$. 
Instead, it obtains a (finite) reward $r$ for each action that it takes. 
Through repeated interactions with the environment, the agent seeks to learn a policy $\pi$ in order to maximize an objective $\mathbb{E}_\pi[\sum_t \gamma^t r(s_t,a_t)]$ \cite{sutton2018reinforcement}. 
%
Let \(\boldsymbol{\pi}\) represent the set of stationary policies of the agent. A policy \(\pi \in \boldsymbol{\pi}\) is considered a deterministic stationary policy if \(\pi \in \{0, 1\}^{|A|}\). It is said to be a stochastic stationary policy if \(\pi \in [0, 1]^{|A|}\).
%


\subsection{Stochastic Approximation (SA) Algorithms}
Let \( h: \mathcal{R}^{m_x} \rightarrow \mathcal{R}^{m_x} \) be a continuous function of 
parameters \( x \in \mathcal{R}^{m_x} \). SA algorithms are designed to solve equations of the form \( h(x) = 0 \) based on noisy measurements of \( h(x) \). The classical SA algorithm is given by:
\begin{equation}\label{eq:SA_basic}
x^{t+1} = x^{t} + \delta^{t}_{x}[h(x^{t}) + n_{x}^{t}], \text{ for } t \geq 0.
\end{equation}
Here, \( t \) denotes the iteration index, and \( x^{t} \) represents the estimate of \( x \) at the \( t^{\text{th}} \) iteration. The term \( n_{x}^{t} \) represents the zero-mean measurement noise associated with \( x^{t} \), and \( \delta^{t}_{x} \) denotes the learning rate. The stationary points of Eq.~\eqref{eq:SA_basic} coincide with the solutions of \( h(x) = 0 \) in the absence of noise (\( n_{x}^{t} = 0 \)). The convergence of SA algorithms is typically analyzed through their associated Ordinary Differential Equations (ODEs), represented as \( \dot{x} = h(x) \).

The convergence of an SA algorithm requires specific assumptions on the learning rate \( \delta^{t}_{x} \):

\begin{assumption}\label{assmp:step-size}
 \(\sum_{t= 0}^{\infty} \delta^{t}_{x} = \infty\) and \(\sum_{t= 0}^{\infty} (\delta^{t}_{x})^{2} < \infty\).
\end{assumption}

Examples of \( \delta^{t}_{x} \) satisfying Assumption~\ref{assmp:step-size} include \( \delta^{t}_{x} = 1/t \) and \( \delta^{t}_{x} = 1/(t\log(t)) \). A general convergence result for SA algorithms is stated below.

\begin{prop}[\cite{kushner2012stochastic,metivier1984applications}]\label{prop:KC_Lemma}
	Consider an SA algorithm in the following form defined over a set of parameters $x \in \mathcal{R}^{m_x}$ and a continuous function  $h: \mathcal{R}^{m_x} \rightarrow \mathcal{R}^{m_x}$.
	\begin{equation}\label{eq:KC_itr}
	x^{t+1} = \Theta({x^{t} + \delta^{t}_{x}[h(x^{t}) + n_{x}^{t} + \kappa^{t}]}),~\text{for $t \geq 0$},
	\end{equation} 
	where $\Theta$ is a projection operator that projects each $x^{t}$ iterate onto a compact and convex set $\Lambda \in \mathcal{R}^{m_x}$ and $\kappa^{t}$ is a bounded random sequence. Let the ODE associated with 
 Eqn.~\eqref{eq:KC_itr} be  
	\begin{equation}\label{eq:KC_ODE}
	\dot{x} = \bar{\Theta}(h(x)), 
	\end{equation}
	where $\bar{\Theta}(h(x)) = \lim\limits_{\eta \rightarrow 0} \frac{\Theta(x + \eta h(x)) - x}{\eta}$ and $\bar{\Theta}$ denotes a projection operator that restricts evolution of Eqn.~\eqref{eq:KC_ODE} to the set $\Lambda$. Let the nonempty compact set $\boldmath{X}$ be a set of asymptotically stable equilibrium points of  Eqn.~\eqref{eq:KC_ODE}.
Then $x^{t}$ converges almost surely to a point in $\boldmath{X}$ as $n \rightarrow \infty$ if 
	\begin{enumerate}
		\item[I)] $\delta^{t}_{x}$ satisfies the conditions in Assumption~\ref{assmp:step-size}.
		\item[II)] $\lim\limits_{t \rightarrow \infty} \left(\sup\limits_{\bar{t} > t}\left|\sum\limits_{l = t}^{\bar{t}}\delta^{t}_{x}n_{x}^{t}\right|\right)  = 0$ almost surely.
		\vspace*{1mm}
		\item[III)] $\lim\limits_{t \rightarrow \infty}\kappa^{t} = 0$ almost surely.
	\end{enumerate}
\end{prop}

\section{The Multi-Critic Actor Critic}\label{sec:ProblemSetup}

In this section we introduce and describe the design of our multi-critic actor-critic (MCAC) algorithm to enable fast autonomy transfer in RL using pre-trained critics. 
To aid readability, we provide a description of notations in Table \ref{table:notation}.

\begin{table}[!h]
\caption{Descriptions of the notations}
\label{table:notation}
\centering
\begin{tabular}{c l}
\hline
\textbf{Notation} & \textbf{Description} \\
\hline
$S$ & State space of the current simulated environment \\
$A$ & Action space of the trained agent \\
$\hat{P}$ & Transition probability kernel of the current simulated env \\
$\hat{r}$ & Reward function of the trained agent\\
$\hat{\pi}$ & Set of stochastic stationary policies for the trained agent \\
$\hat{\pi}(s)$ & Stochastic stationary policy at state $s$ \\
 & for the trained agent \\
$\hat{P}(s'|s, a)$ & Probability of transitioning from state $s$ to $s'$ \\
& under action $a$ in the current simulated environment \\
$\hat{r}(s, a, s')$ & Reward received when transitioning from state $s$ to $s'$ \\
 &  under action $a$ for the trained actor \\
$\hat{V}$ & Modulated value vector of the trained actor \\
$\hat{V}(s)$ & Modulated value function of the trained actor at state $s$ \\
$\bar{V}_{i}$ & Value vector of the $i$-th pre-trained critic \\
$\bar{V}_{i}(s)$ & Value function of the $i$-th pre-trained critic at state $s$ \\
$W$ & Modulation weight vector \\
$\gamma$ & Discount factor \\
$\delta_{W}^{t}$ & Learning rate for weight updates at iteration $t$ \\
$\Theta_W$ & Projection operator for weight updates \\
$\delta_{\hat{\pi}}^{t}$ & Learning rate for policy updates of the trained actor \\
& at iteration $t$ \\
$\Theta_{\hat{\pi}}$ & Projection operator for policy updates of trained actor \\
\hline
\end{tabular}
\end{table}

Our intuition for the MCAC algorithm is grounded in the principles of ensemble learning in machine learning \cite{dong2020survey} and model soups in large language models \cite{wortsman2022model}. These solutions utilize multiple trained models to achieve better performance with significantly fewer resources and and in shorter time. 
Following a similar line of thought, we hypothesize that using multiple pre-trained critic values from training environments under different conditions (e.g., different obstacle orientations) can enhance performance and achieve faster convergence in actor-critic algorithms trained for a new environment with varied conditions. 

Specifically, given access to a set of pretrained critic value functions, the MCAC approach aims to bypass the process of training a critic for the new environment from scratch, opting instead to use a weighted sum of the existing knowledge from the pretrained critic values for estimating the value function in the current environment. 
A weighted average of pre-trained critic value functions will provide a better initial condition for the value function estimated for the new environment. This, in turn, will improve the agent's exploration and achieve better performance in terms of the accumulated reward at convergence. 
Further, since MCAC will not require training new value functions, but rather focuses on finding the best set of weights to compute the weighted average of the pre-trained value functions of critics to estimate the value function of the current environment, it reduces the number of trainable parameters related to critics from $|S|$ to $N$, where $|S|$ is the cardinality of the state space of the environment, and $N$ is the number of pre-trained critic value functions available in the new environment. Typically, $N \ll |S|$, which aids the agent in achieving significant speedup in training. Additionally, starting from a better initial condition, as mentioned previously, will also help the agent learn faster compared to learning from scratch.



In order to setup the MCAC approach and algorithm, we first establish an assumption regarding the availability of pre-trained value functions, derived using the AC algorithm from various environment settings. 


Let $\bar{V}_{i}(s)$ be the value function of the $i^{\text{th}}$ pre-trained critic at state $s \in {S}$. Then define the vector $\bar{V}_i := [\bar{V}_{i}(s)]_{s \in {S}}$.

\begin{assumption}\label{assmp:pre-trained-cv}
    Converged value vectors $\bar{V}_i$ for $i = 1, 2, \ldots, N$ pre-trained critics are readily available for the training the actor (i.e., agent) using the MCAC algorithm.
\end{assumption}

Let $W = [w_{i}]_{i=1}^{N}$  be the weight vector where each weight $w_{i}$ is associated with a pre-trained value function $\bar{V}_i$ of critic $i$. Further, let $\sum_{i=1}^{N} w_{i} = 1$ and $w_{i} \geq 0$ for all $i\in \{1, 2, \ldots, N\}$. Then, the value vector $\hat{V} = [\hat{V}(s)]_{s \in {S}}$ of the actor is defined by
\begin{equation}\label{eq:mod-val}
    \hat{V}  = \Big[\sum_{i=1}^{N}w_{i}\bar{V}_{i}(s)\Big]_{s \in S}.
\end{equation}

The temporal difference (TD) error associated with the trained actor at state $s \in {S}$ when taking action $a \in {A}$ is: 
\begin{eqnarray}
   \sum_{a \in {A}}\Big[\hat{V}(s) - \gamma\sum_{s' \in {S}}\hat{P}(s,a,s')[\hat{V}(s') - \hat{r}(s,a,s')]\Big]\hat{\pi}(s,a),\nonumber
\end{eqnarray}
where $0< \gamma < 1$ denotes the discount factor.
In actor-critic (AC) RL algorithms \cite{sutton2018reinforcement}, the TD error represents the difference between the predicted  reward 
and the actual reward 
obtained. The TD error is used to update the value function and the policy in AC algorithms. A high TD error indicates that the predictions are quite different from the actual outcomes, suggesting that the actor's value function and the policy need significant updating. Conversely, a low TD error suggests that the actor's predictions are accurate.

Minimizing the TD error for the trained actor can be expressed as the following optimization problem:


\begin{align}
&\min_{w, \hat{\pi}} \sum_{a \in {A}}\Big[\hat{V}(s) - \gamma\sum_{s' \in {S}}\hat{P}(s,a,s')[\hat{V}(s') - \hat{r}(s,a,s')]\Big] \nonumber \\ 
& \qquad \qquad \qquad \qquad \qquad \qquad \qquad \qquad \qquad \qquad \hat{\pi}(s,a) \nonumber \\
    &\text{subject to:} \nonumber \\
    &\hat{V}(s) - \gamma\sum_{s' \in {S}}\hat{P}(s,a,s')[\hat{V}(s') - \hat{r}(s,a,s')] \geq 0, \label{C1}\\
    &\qquad \qquad \qquad \qquad \qquad \forall s, s' \in {S}, { a \in {A}}, \nonumber  \\
    &  \hat{V}(s)  = \sum_{i=1}^{N}w_{i}\bar{V}_{i}(s), \quad \forall s \in {S} \label{C2}\\ 
    & \sum_{a \in {A}}\hat{\pi}(s,a) = 1, \quad \forall s, s' \in {S}   \label{C3}\\
    & \hat{\pi}(s,a) \geq 0, \quad \forall s \in {S} {\text{ and } a \in {A}} \label{C4}
\end{align}


\begin{remark}
If the above optimization problem was framed within a typical actor-critic (AC) framework, then the minimization would be performed over \( \hat{V} \) and \( \hat{\pi} \), while omitting the equality constraint in Eqn. \( \eqref{C2} \). Then, using a stochastic approximation algorithm to solve the standard AC algorithm would provide the following critic recursion.
\begin{equation*}
    \hat{V}^{t+1}(s) = \hat{V}^{t}(s) + \delta^{t}_{V}[\gamma\hat{V}^{t}(s') - \hat{V}^{t}(s) + \hat{r}(s,a,s')].
\end{equation*}
Here $\delta^{t}_{V}$ is the learning rate of the value updates.
\end{remark}



In our MCAC algorithm, we replace the AC critic update by weight updates in the following way:
\begin{equation}
     w_i^{t+1} = w_i^{t} + \delta^{t}_{W}\sqrt{w_i^t}[\gamma\bar{V}_{i}(s') - \bar{V}_{i}(s) + \hat{r}(s,a,s')]. \nonumber 
\end{equation}
Here the term $\gamma\bar{V}_{i}(s') - \bar{V}_{i}(s) + \hat{r}(s,a,s')$ ensures weights are updated in a descent direction of the estimated TD error with respect to the pre-trained critic value functions and $\delta^{t}_{W}$ is the learning rate of the weight updates.
We further limit evolution of the weight vector $W^{t}$ to a probability simplex defined by  $\sum_{i=1}^{N} w_{i}^{t} = 1$ and $w_{i}^{t} \geq 0$ for all $i\in \{1, 2, \ldots, N\}$. Therefore, we modify above weight updates by introducing a projection map $\Theta_{W}$ as follows:
\begin{equation}
     w_i^{t+1} = {\Theta}_{W}(w_i^{t} + \delta^{t}_{W}\sqrt{w_i^t}[\gamma\bar{V}_{i}(s') - \bar{V}_{i}(s) + \hat{r}(s,a,s')]). \label{eq:extended_weight}
\end{equation}

The rewards $\hat{r}_i(s,a,s')$ associated with environments of the $i^{\text{th}}$ pre-trained critic are not directly accessible during the actor's training phase during the MCAC algorithm. Hence, availability of these rewards may be limited. We propose a possible methodology for approximating values of $\hat{r}_i(s,a,s')$ in instances where they are not explicitly known below.


Following the standard definitions of the policy updates for actor-critic reinforcement learning algorithm, we get:
\begin{equation}\label{eq:extended_policy}
    \hat{\pi}^{t+1}(s,a)  \hspace{-1mm} =  \hspace{-1mm}{\Theta}_{\pi}(\hat{\pi}^{t}(s,a) \hspace{-0.5mm} - \hspace{-0.5mm} \delta_{\pi}^{t}\sqrt{\pi^{t}(s,a)} \hat{E}^{t}(s,a,s')). \nonumber 
\end{equation}

Here $\hat{E}^{t}(s,a,s') = \sum_{i=1}^{m}w_{i}^t[\bar{V}_{i}^t(s) - \gamma\bar{V}_{i}^t(s')] - \hat{r}(s,a,s')$ ensures that policies are updated in a descent direction of the total estimated TD error computed using the weighted pre-trained critic value functions. The projection map $\Theta_{\pi}$ ensures that evolution of the policy is such that $\sum_{a \in A(s)} \pi(s,a) = 1$ and $\pi(s,a) \geq 0$. $\delta_{\pi}^{t}$ is the learning rate of policy updates and is chosen such that $\delta_{\pi}^{t} \ll \delta_{W}^{t}$ for all $t$. 
Thus, the policy is updated in a slower time-scale compared to weights. 
Due to time scale separation, iterations in faster time scales see iterations in slower times scales as quasi-static, while the latter sees former as nearly equilibriated (Chap. 6 of \cite{borkar2009stochastic}). 

The setup described in this section leads to design of our MCAC Algorithm, shown in Algorithm \ref{algo}. 
Next, we present results to establish convergence of weight update iterations in the MCAC Algorithm to an equilibrium, and convergence of policy update iterations to a stable equilibrium point. 
We will leverage Proposition~\ref{prop:KC_Lemma} to establish convergence of the updates. 
We defer detailed proofs of these results to an extended version of the paper. 



\begin{center}
	\begin{algorithm}[!h]
		\caption{MCAC Algorithm}
		\label{algo:Stack}
		\begin{algorithmic}[1]
			\State \textbf{Input:} State space (${{S}}$), Rewards (r), Pre-trained value vectors $\bar{V}_i$ for $i = 1, 2, \ldots, m$, Rewards of pre-trained actors $\bar{r}_i$ for $i = 1, 2, \ldots, N$, Max. iterations ($T >> 0$)
			\State \textbf{Output:} Policy, $\hat{\pi}^\* \leftarrow \hat{\pi}^{T}$
			\State \textbf{Initialization: } $t \leftarrow 0$, $W^{0} \leftarrow \frac{1}{N}\mathbf{1}$, $\hat{\pi}^{0}~\leftarrow~\boldsymbol{\hat{\pi}}$,  $s \leftarrow s_0$
			\While  {$t \leqslant T$} 
			\State Draw $a$ from $\hat{\pi}(s)$
			\State Reveal the next state $s'$ according to ${P}$
			\State Observe the reward $\hat{r}(s,a,s')$ 
            \State $\hat{E}^{t}_i(s,a,s') = \bar{V}_i(s) - \gamma\bar{V}_i(s') + \hat{r}_i(s,a,s')$ 
			\State   $w_i^{t+1} = {\Theta}_{W}(w_i^{t} + \delta^{t}_{W}\sqrt{w_i}\hat{E}^{t}_i(s,a,s'))$ \label{eq:w_itr}
            \State  $\hat{E}^{t}(s,a,s') = \sum_{i=1}^{m}w_{i}^t[\bar{V}_{i}(s) - \gamma\bar{V}_{i}(s')] + \hat{r}(s,a,s')$
			\State  $\hat{\pi}^{t+1}(s,a)  \hspace{-1mm} =  \hspace{-1mm}{\Theta}_{\hat{\pi}}(\hat{\pi}^{t}(s,a) \hspace{-0.5mm} - \hspace{-0.5mm} \delta_{\pi}^{t}\sqrt{\hat{\pi}^{t}(s,a)} \hat{E}^{t}(s,a,s'))$
			\State Update the state: $s \leftarrow s'$
			\State $ t \leftarrow t + 1$
			\EndWhile 
		\end{algorithmic}\label{algo}
	\end{algorithm}
	\vspace*{-3 mm}
\end{center}



\begin{figure*}[ht]
    \centering
    \begin{subfigure}[b]{0.32\textwidth}
        \centering
        \includegraphics[width=\textwidth]{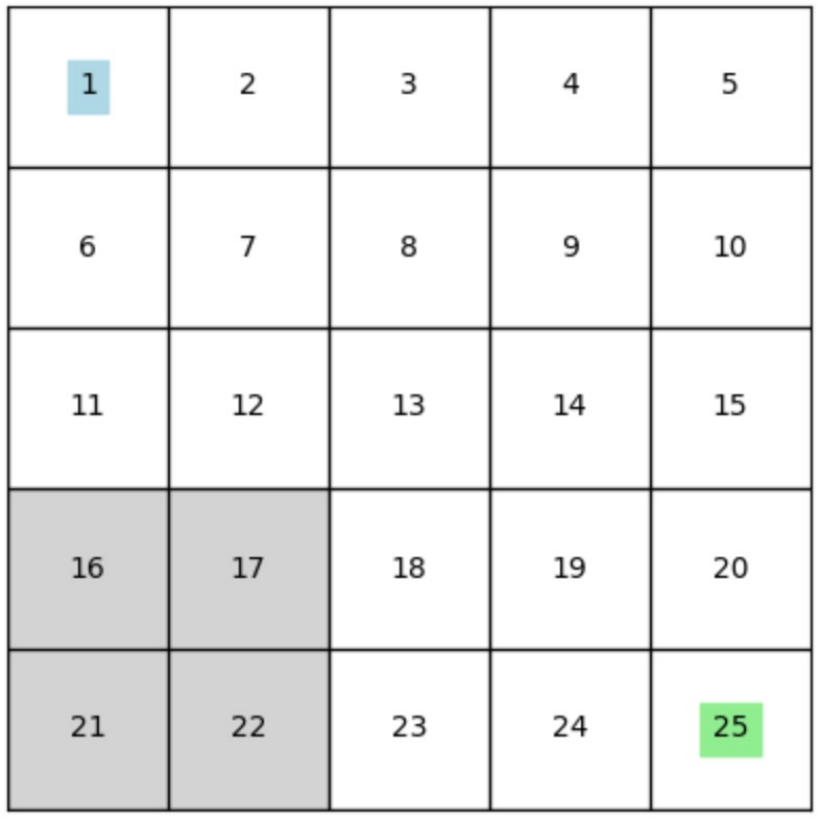}
        \caption{Pre-Trained Scenario 1}
    \end{subfigure}
    \hfill 
    \begin{subfigure}[b]{0.32\textwidth}
        \centering
        \includegraphics[width=\textwidth]{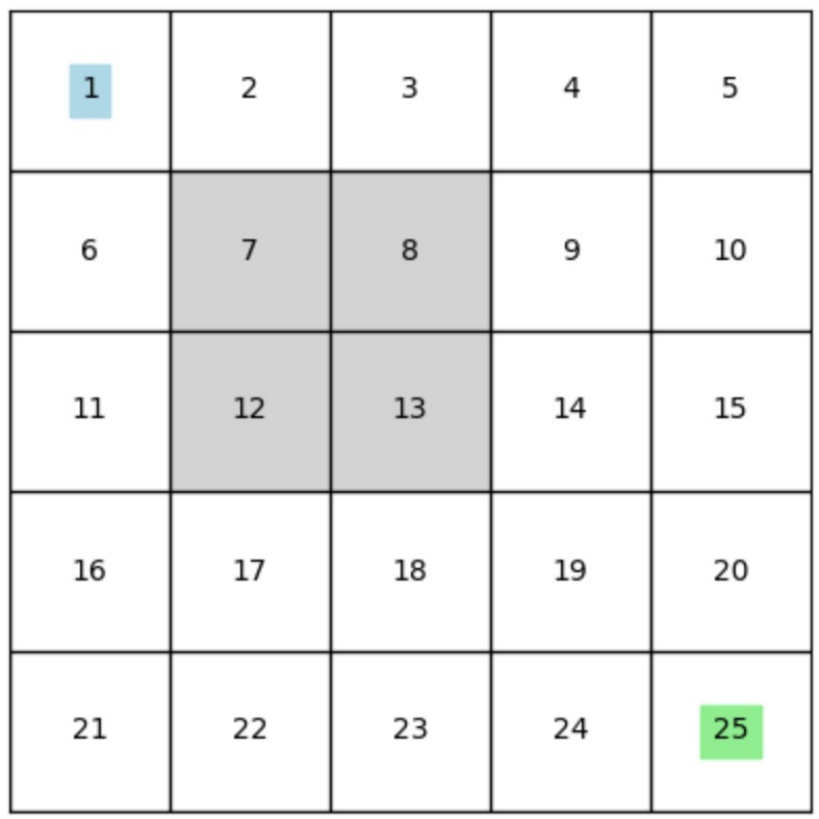}
        \caption{Pre-Trained Scenario 2}
    \end{subfigure}
    \hfill
    \begin{subfigure}[b]{0.32\textwidth}
        \centering
        \includegraphics[width=\textwidth]{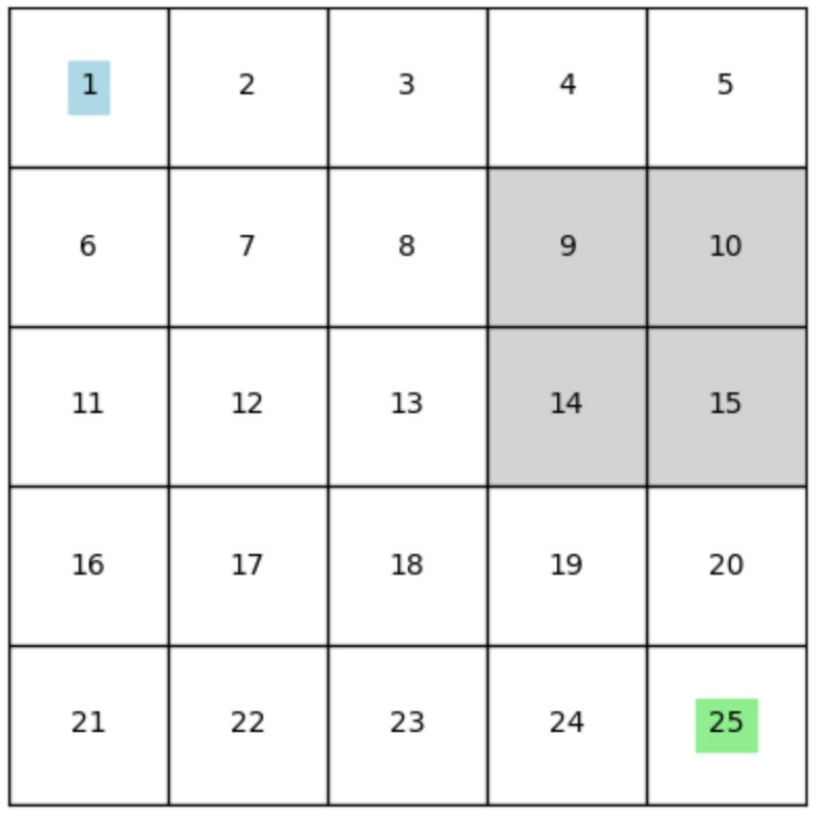}
        \caption{Pre-Trained Scenario 3}
    \end{subfigure}

    \bigskip 

    \begin{subfigure}[b]{0.24\textwidth}
        \centering
        \includegraphics[width=\textwidth]{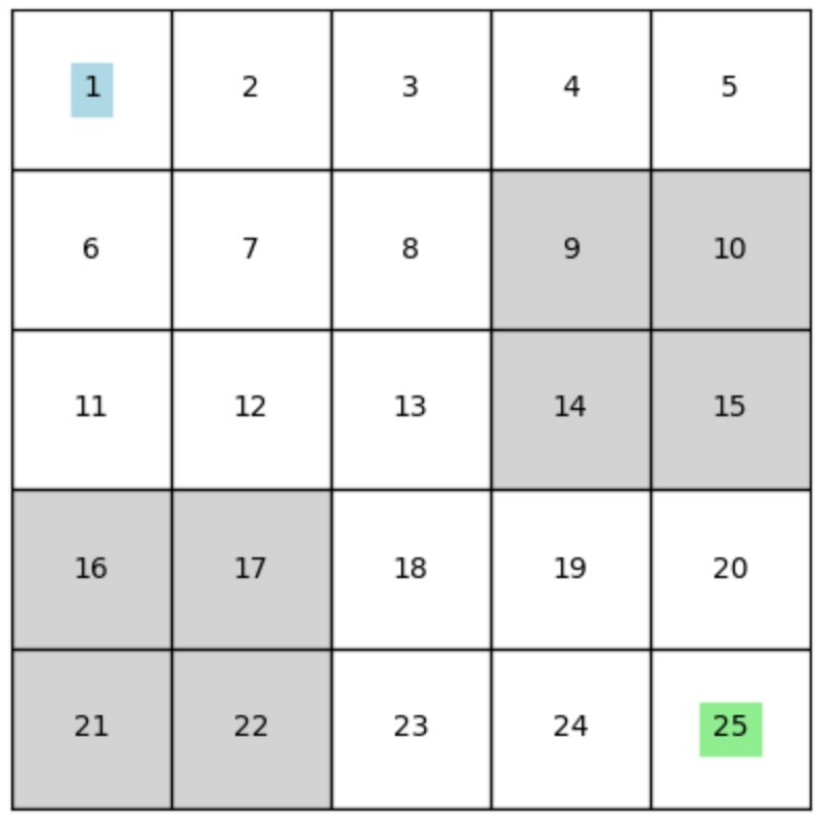}
        \caption{Deployment Scenario 1}
    \end{subfigure}
    \hfill
    \begin{subfigure}[b]{0.24\textwidth}
        \centering
        \includegraphics[width=\textwidth]{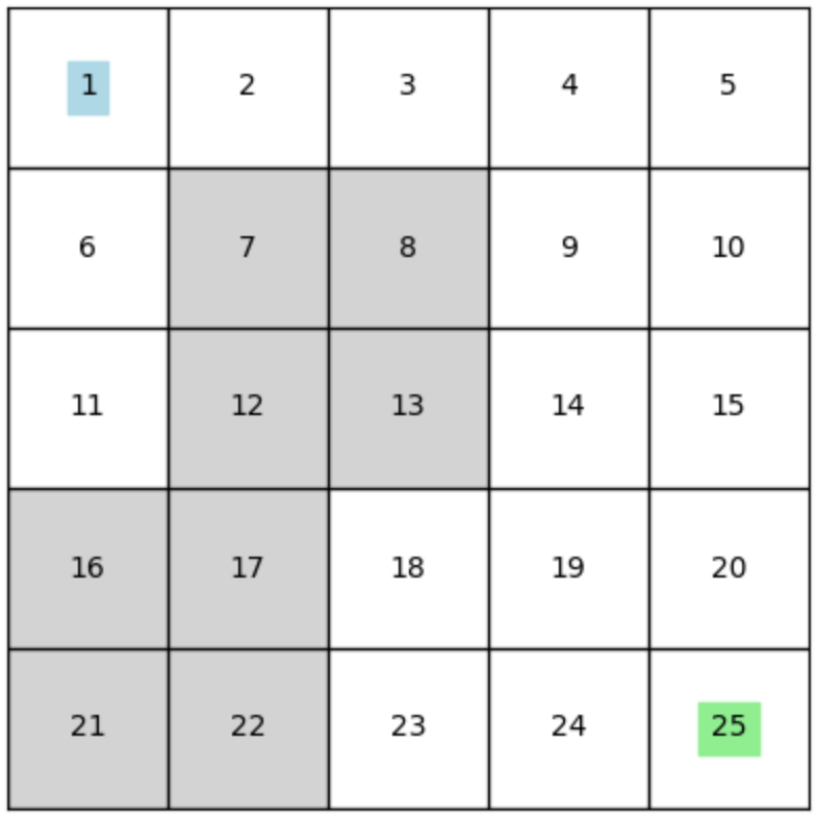}
        \caption{Deployment Scenario 2}
    \end{subfigure}
    \hfill
    \begin{subfigure}[b]{0.24\textwidth}
        \centering
        \includegraphics[width=\textwidth]{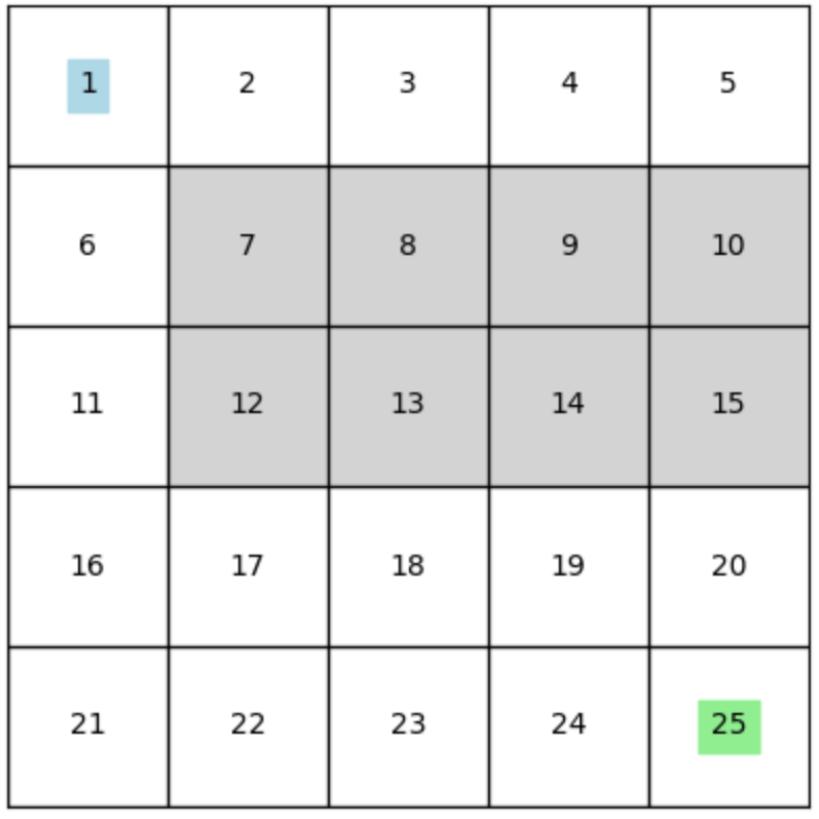}
        \caption{Deployment Scenario 3}
    \end{subfigure}
    \hfill
    \begin{subfigure}[b]{0.24\textwidth}
        \centering
        \includegraphics[width=\textwidth]{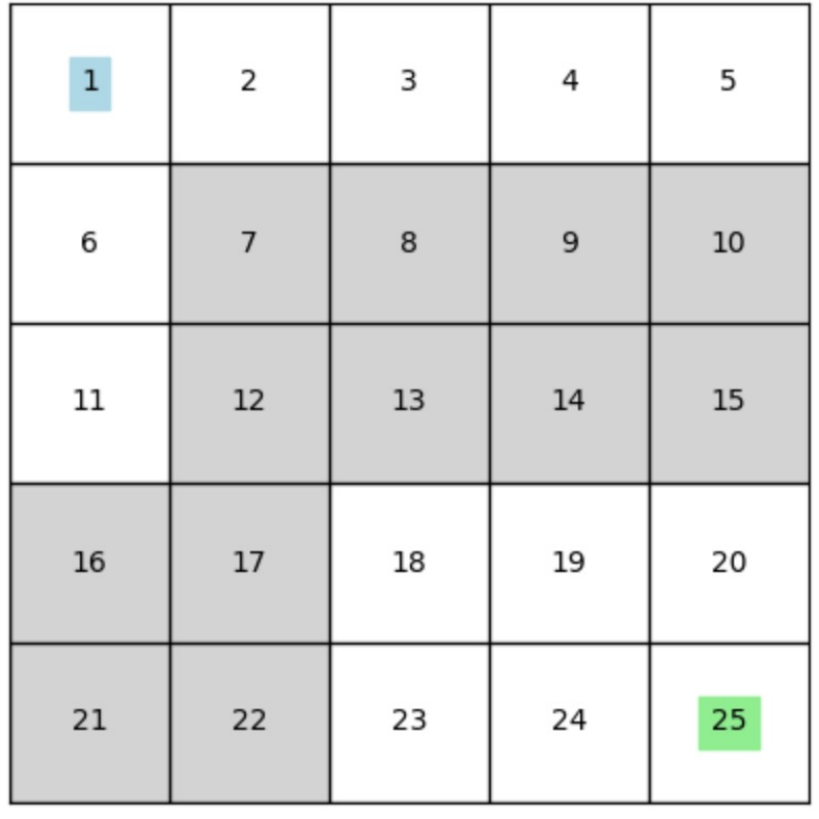}
        \caption{Deployment Scenario 4}
    \end{subfigure}
    \caption{\textbf{Case Study 1 Setup}: This figure shows a $5 \times 5$ grid. The blue (green) color state is the start (goal) state and the grey shaded states represent obstacles. The top row shows the configurations of obstacles that are used to obtain the pretrained critics. The bottom row shows the obstacle configurations on which our MCAC algorithm is evaluated. The four deployment scenarios in the bottom row are obtained by combining the pretrained scenarios in the top row in different ways.}\label{fig:SetUp1}
\end{figure*}



\begin{theorem}\label{thm:policy-convergence}
 The weight updates $w^{t}_{i}$ for all $i \in \{1, 2, \ldots, N\}$ in Line~9 of Algorithm~\ref{algo} converge to a stable equilibrium point $W^{\star} \in \Omega$.
\end{theorem}

\begin{theorem}\label{thm:policy-convergence}
 The policy updates $\pi^{t}(s,a)$ for all $a \in \A(s)$, $s \in \mathbf{S}$ in Line~11 of Algorithm~\ref{algo} converge to a stable equilibrium point $\pi^{\star} \in \Pi$.
\end{theorem}

\section{Experiments}\label{sec:Experiments}

\begin{figure*}[ht]
    \centering
    \begin{subfigure}[b]{0.19\textwidth}
        \centering
        \includegraphics[width=\textwidth]{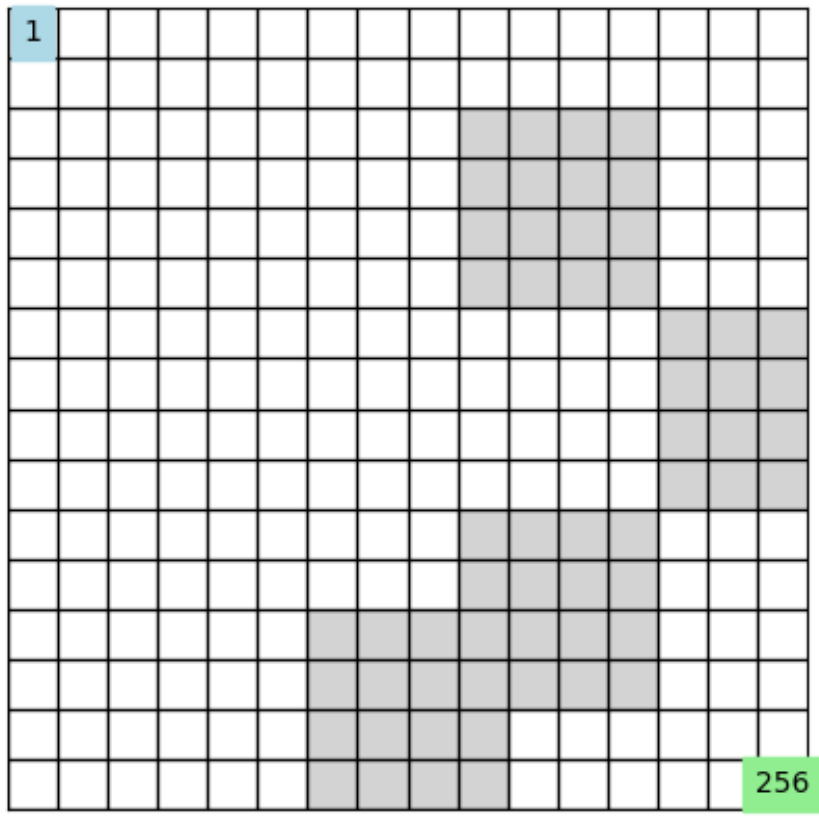}
        \caption{Pre-Trained Scenario 1}
    \end{subfigure}
    \hfill 
    \begin{subfigure}[b]{0.19\textwidth}
        \centering
        \includegraphics[width=\textwidth]{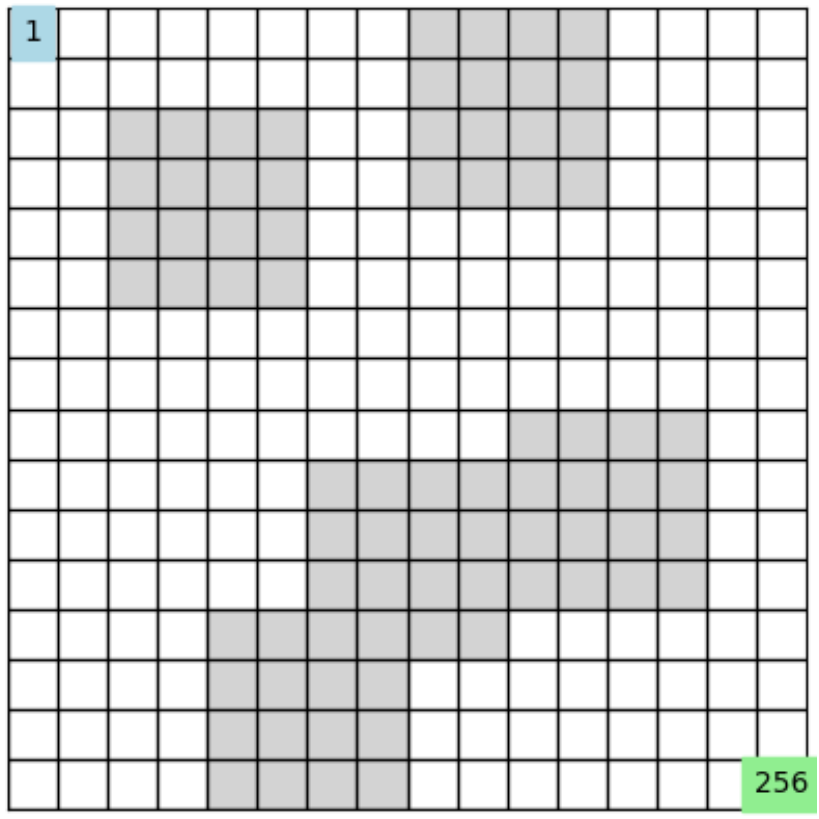}
        \caption{Pre-Trained Scenario 2}
    \end{subfigure}
    \hfill
    \begin{subfigure}[b]{0.19\textwidth}
        \centering
        \includegraphics[width=\textwidth]{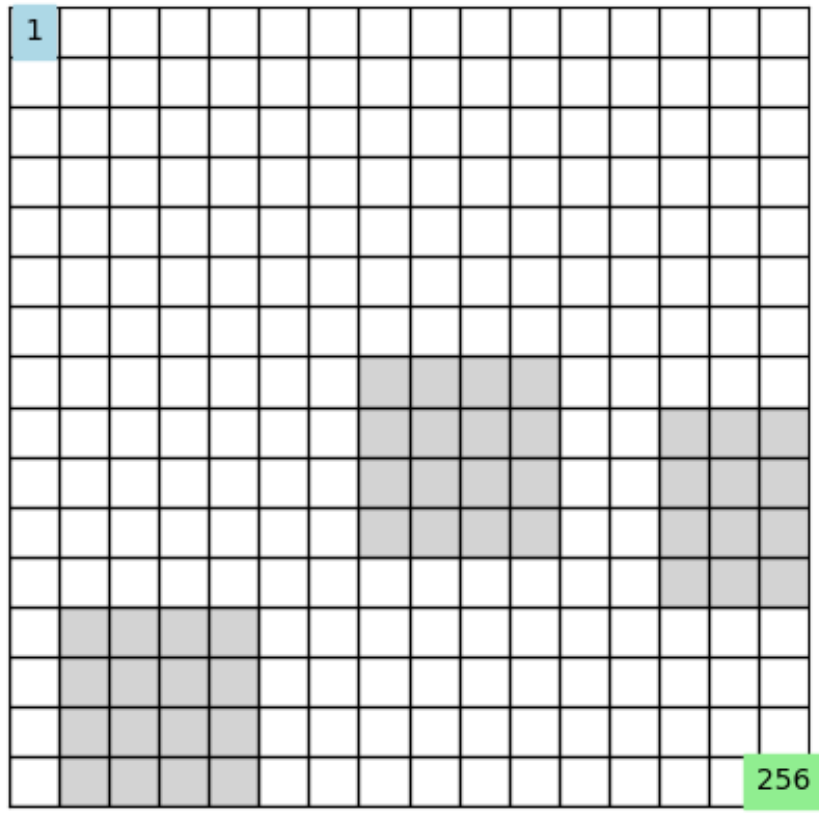}
        \caption{Pre-Trained Scenario 3}
    \end{subfigure}
    \hfill
    \begin{subfigure}[b]{0.19\textwidth}
        \centering
        \includegraphics[width=\textwidth]{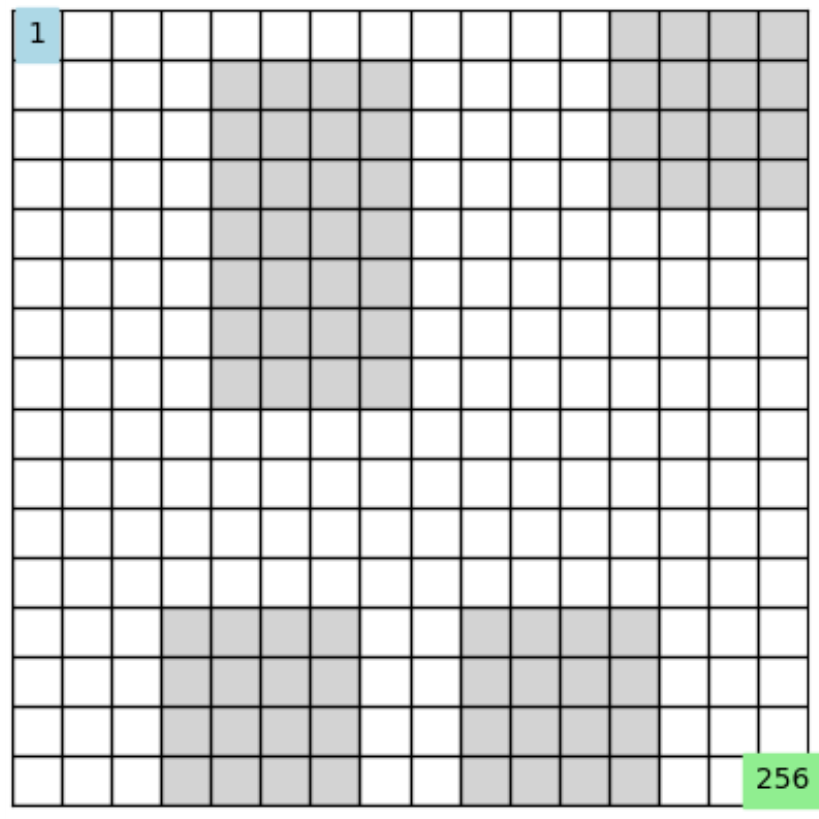}
        \caption{Pre-Trained Scenario 4}
    \end{subfigure}
    \hfill
    \begin{subfigure}[b]{0.19\textwidth}
        \centering
        \includegraphics[width=\textwidth]{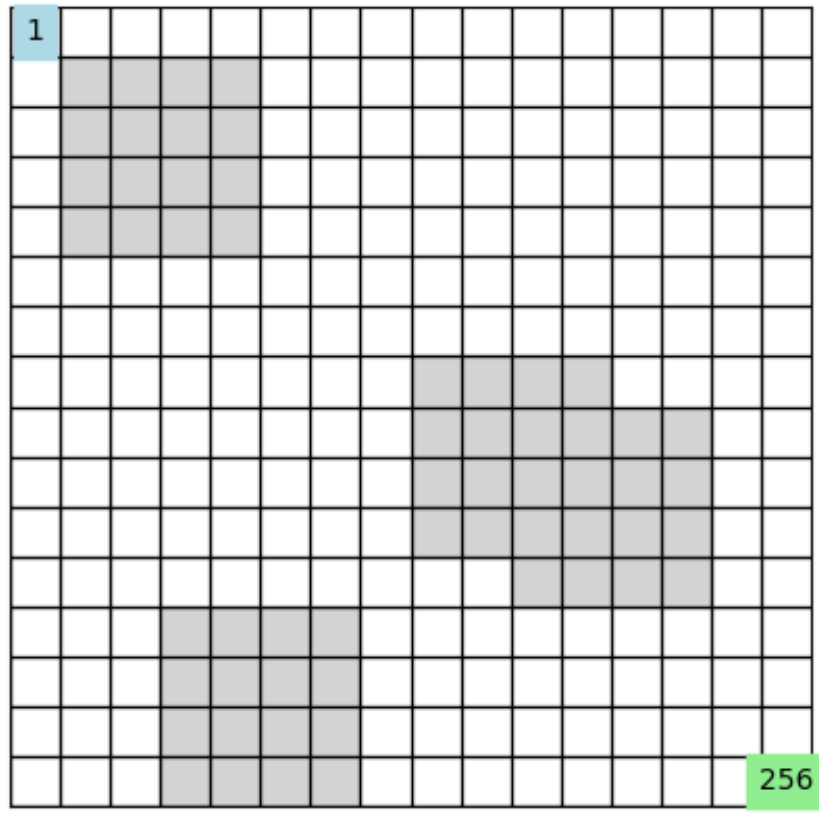}
        \caption{Pre-Trained Scenario 5}
    \end{subfigure}

    \bigskip 

    \begin{subfigure}[b]{0.24\textwidth}
        \centering
        \includegraphics[width=\textwidth]{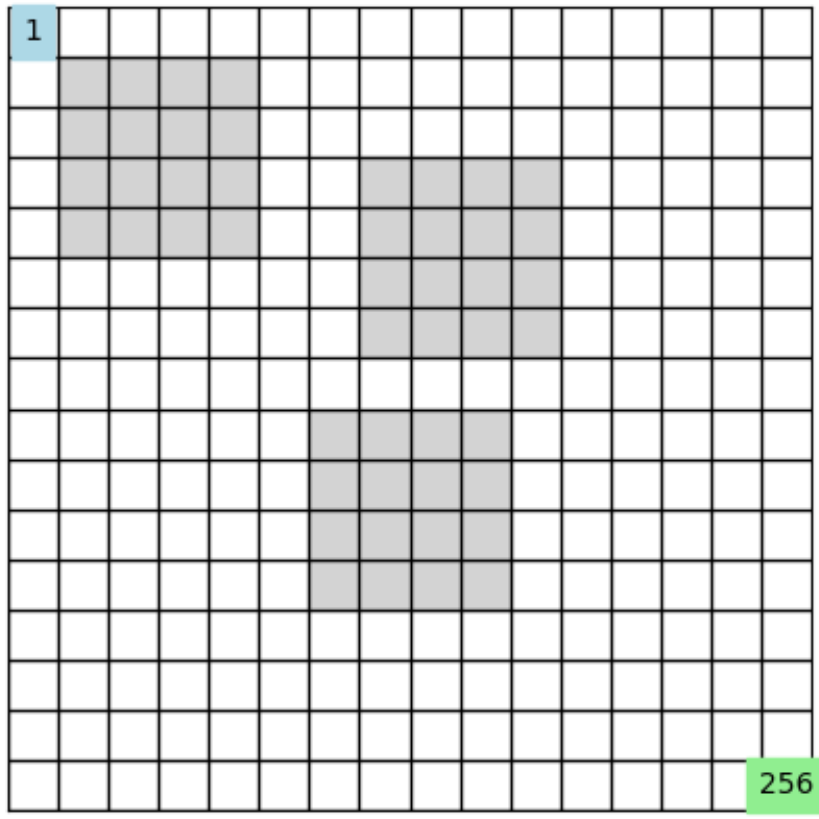}
        \caption{Deployment Scenario 1}
    \end{subfigure}
    \hfill
    \begin{subfigure}[b]{0.24\textwidth}
        \centering
        \includegraphics[width=\textwidth]{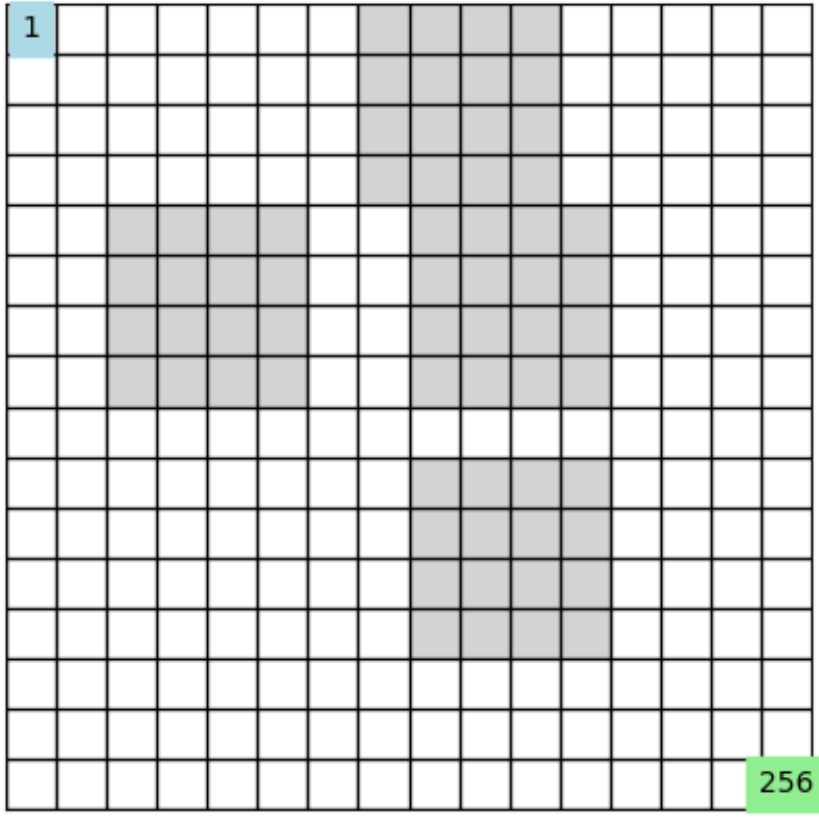}
        \caption{Deployment Scenario 2}
    \end{subfigure}
    \hfill
    \begin{subfigure}[b]{0.24\textwidth}
        \centering
        \includegraphics[width=\textwidth]{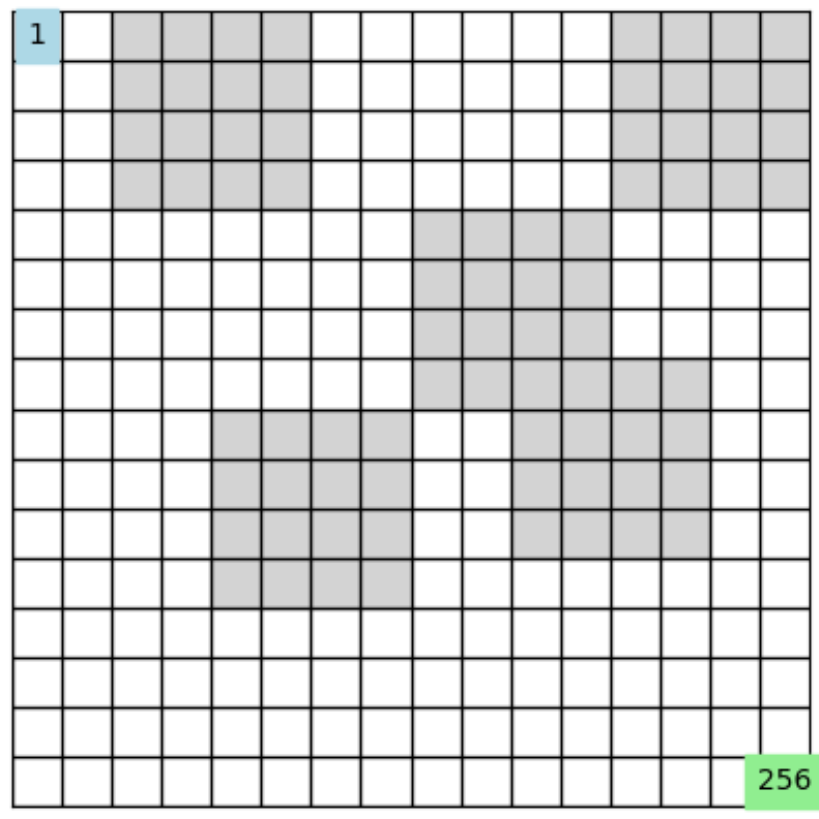}
        \caption{Deployment Scenario 3}
    \end{subfigure}
    \hfill
    \begin{subfigure}[b]{0.24\textwidth}
        \centering
        \includegraphics[width=\textwidth]{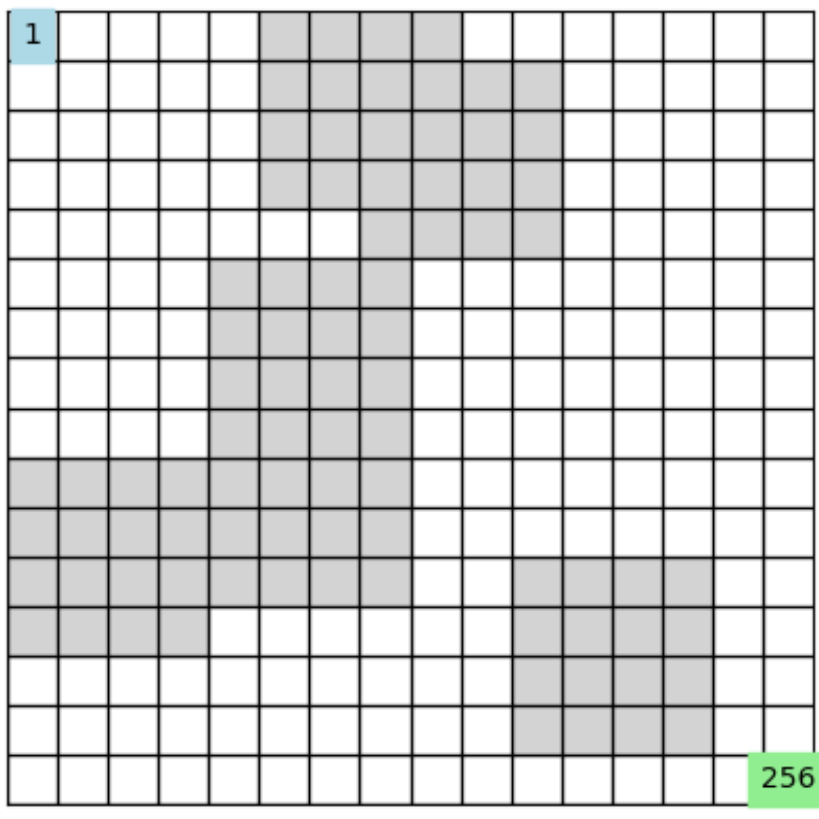}
        \caption{Deployment Scenario 4}
    \end{subfigure}
    \caption{\textbf{Case Study 2 Setup}: This figure shows a $16 \times 16$ grid. The blue (green) color state is the start (goal) state and the grey shaded states represent obstacles. The top row shows the configurations of obstacles that are used to obtain the pretrained critics. The bottom row shows the obstacle configurations on which our MCAC algorithm is evaluated. The four deployment scenarios in the bottom row are obtained by combining the pretrained scenarios in the top row in different ways.}\label{fig:SetUp2}
\end{figure*}

This section presents two case studies to compare our proposed multi-critic actor-critic (MCAC) algorithm against a baseline actor-critic (Baseline AC).
We first present the details of the two case study scenarios developed for comparing the two algorithms. 
Next we describe the hyper parameters used in Baseline AC and MCAC algorithms and define metrics used to compare both algorithms. 
We then present a comparison of results from the Baseline AC and our MCAC algorithm on two case study scenarios.

\subsection{Case Study Environments}
We demonstrate effectiveness of our MCAC algorithm through experiments conducted in two distinct grid world environments of dimensions $5\times5$ ({\bf Case Study 1}) and $16\times16$ ({\bf Case Study 2}). Our experiments evaluate capabilities of RL agents in scenarios of varying complexity. Each grid world is populated with obstacles and potentially hazardous areas, challenging the agent to learn to navigate effectively to a goal location, starting from an initial position. 
In the $5 \times 5$ grid, $1 \times 1$ obstacles are placed at predefined grid locations; the $16 \times 16$ grid features a more complex arrangement of obstacles, with $4 \times 4$ obstacles placed at predefined locations. 
Details about the grid environments are presented below. 


\noindent{\bf State Space:}
The state space, ${S}$, of our grid-worlds is defined by the discrete positions that an agent can occupy on a grid. For a grid of size $L \times L$, the state space has $L^2$ possible positions, each uniquely identified by its grid coordinates $(x, y)$. We examine $5 \times 5$ and $16 \times 16$, grids leading to state spaces of $25$ and $256$ distinct states, respectively.

\noindent{\bf Action Space:}
The action space, $A$, available to the agent consists of four deterministic actions: move up ($\uparrow$), move down ($\downarrow$), move left ($\leftarrow$), and move right ($\rightarrow$). When the agent takes an action in a certain grid position (state), it moves one position in that direction, unless there is something in the way (e.g., obstacles, edge of grid). When the agent attempts to move beyond the grid's edges, 
the agent's position remains unchanged. 

\noindent{\bf Transition Probabilities: } The transition probabilities, $P(s'|s, a)$, describe the likelihood of moving from state $s$ to state $s'$ given action $a$. In our grid-worlds, if there is no obstacle, these probabilities are either $0$ or $1$; a probability of $1$ indicates that action $a$ taken in state $s$ will always result in transition to state $s'$, and $0$ otherwise. When the agent is in an obstacle state, the agent will stay in the obstacle state with probability $p_{\text{Stay}} = 0.75$ and move to the state suggested by the action  with probability $1 - p_{\text{Stay}} = 0.25$.

\noindent{{\bf Reward Structure}}
The agent receives reward $r(s, a, s')$ for each transition from state $s$ to state $s'$ via action $a$. Specifically, $r(s, a, s') = 100$ when the agent takes an action that reaches the goal state, and $r(s, a, s')=-1$ for every other action taken by the agent. 
This reward structure aims to balance exploration and exploitation by incentivizing the discovery of the shortest path to the goal. 

\subsection{Experiment Setup}

\begin{table*}[!ht]
\caption{This table presents a comparison of the actor-critic (AC) algorithm and our multi-critic actor critic (MCAC) algorithm on {\bf Case Study 1} using four metrics: average total reward at convergence ($\uparrow$), average number of steps to reach goal state from start state ($\downarrow$), average runtime to convergence ($\downarrow$), and number of episodes to convergence ($\downarrow$). All averages are calculated over 100 independent runs. Our results show that across all deployment scenarios, MCAC consistently achieves higher average reward, converges in fewer steps, takes less time to convergence (with speedup (SU1) up to \textcolor{black}{2.31x}), and fewer episodes to convergence (with speedup (SU2) up to \textcolor{black}{10.44x}). The total speedup $SU = SU1 \times SU2$ of our MCAC algorithm is up to \textbf{\textcolor{black}{22.76x}} compared to the baseline AC algorithm and \textbf{\textcolor{black}{14.35x}} on average.}\label{table:ResultsCase1}
\centering
\begin{tabular}{|c|c|c|c|c|c|c|c|c|c|c|c|}
\hline
\multirow{3}{*}{\parbox{2cm}{\centering \textbf{Deployment\\Scenario}}} & \multicolumn{2}{c|}{\textbf{Average Total Reward}} & \multicolumn{2}{c|}{\textbf{Average Number of Steps}} & \multicolumn{3}{c|}{\textbf{Average Runtime}} & \multicolumn{3}{c|}{\textbf{Number of Episodes}} & \multirow{3}{*}{\textbf{Total SU}} \\ \cline{2-11}
                                    & \textbf{AC} & \textbf{MCAC} & \textbf{AC} & \textbf{MCAC} & \textbf{AC} & \textbf{MCAC} & \textbf{SU1} & \textbf{AC} & \textbf{MCAC} & \textbf{SU2} & \\ \cline{2-11}
                                    & & & & & & & & & & & \\ \hline
\textbf{1}               & $85.98$ & $\mathbf{90.79}$ & 15.02 & $\mathbf{10.21}$ & $0.18$ s & $\mathbf{0.07}$ s & $2.41\times$ & 98 & 26 & $3.77\times$ & $9.09\times$ \\ \hline
\textbf{2}               & $84.25$ & $\mathbf{89.37}$ & 16.75 & $\mathbf{11.63}$ & 0.17 s & $\mathbf{0.08}$ s & $2.18\times$ & 94 & 9 & $10.44\times$ & $22.76\times$ \\ \hline
\textbf{3}            & $82.44$ & $\mathbf{89.07}$ & 18.56 & $\mathbf{11.93}$ & $0.17$ s & $\mathbf{0.08}$ s & $2.17\times$ & 98 & 20 & $4.90\times$ & $10.63\times$ \\ \hline
\textbf{4}            & 80.73 & $\mathbf{87.73}$ & 20.27 & $\mathbf{13.27}$ & $0.20$ s & $\mathbf{0.08}$ s & $2.31\times$ & 97 & 15 & $6.47\times$ & $14.95\times$ \\ \hline
\end{tabular}
\end{table*}

\begin{table*}[!h]
\caption{This table presents a comparison of the actor-critic (AC) algorithm and our multi-critic actor critic (MCAC) algorithm on {\bf Case Study 2} using four metrics: average total reward at convergence ($\uparrow$), average number of steps to reach goal state from start state ($\downarrow$), average runtime to convergence ($\downarrow$), and number of episodes to convergence ($\downarrow$). All averages are calculated over 100 independent runs. Our results show that across all deployment scenarios, MCAC consistently achieves higher average reward, converges in fewer steps, takes less time to convergence (with speedup (SU1) up to \textcolor{black}{2.06x}), and fewer episodes to convergence (with speedup (SU2) up to \textcolor{black}{4.76x}). The total speedup $SU = SU1 \times SU2$ of our MCAC algorithm is up to \textbf{\textcolor{black}{9.37x}} compared to the baseline AC algorithm, and \textbf{\textcolor{black}{8.44x}} on average.}\label{table:ResultsCase2}
\centering
\begin{tabular}{|c|c|c|c|c|c|c|c|c|c|c|c|}
\hline
\multirow{3}{*}{\parbox{2cm}{\centering \textbf{Deployment\\Scenario}}} & \multicolumn{2}{c|}{\textbf{Average Total Reward}} & \multicolumn{2}{c|}{\textbf{Average Number of Steps}} & \multicolumn{3}{c|}{\textbf{Average Runtime}} & \multicolumn{3}{c|}{\textbf{Number of Episodes}} & \multirow{3}{*}{\textbf{Total SU}} \\ \cline{2-11}
                                    & \textbf{AC} & \textbf{MCAC} & \textbf{AC} & \textbf{MCAC} & \textbf{AC} & \textbf{MCAC} & \textbf{SU1} & \textbf{AC} & \textbf{MCAC} & \textbf{SU2} & \\ \cline{2-11}
                                    & & & & & & & & & & & \\ \hline
\textbf{1}               & $-236.95$ & $\mathbf{-70.41}$ & 337.95 & $\mathbf{171.41}$ & $2.66$ s & $\mathbf{1.40}$ s & $1.89\times$ & 98 & 25 & $3.92\times$ & $7.41\times$ \\ \hline
\textbf{2}               & $-223.89$ & $\mathbf{-82.40}$ & 324.89 & $\mathbf{183.40}$ & 2.99 s & $\mathbf{1.45}$ s & $2.06\times$ & 100 & 22 & $4.55\times$ & $9.37\times$ \\ \hline
\textbf{3}            & $-235.33$ & $\mathbf{-90.86}$ & 336.33 & $\mathbf{191.86}$ & $2.85$ s & $\mathbf{1.49}$ s & $1.91\times$ & 100 & 21 & $4.76\times$ & $9.09\times$ \\ \hline
\textbf{4}            & -256.16 & $\mathbf{-109.88}$ & 357.16 & $\mathbf{210.88}$ & $3.00$ s & $\mathbf{1.55}$ s & $1.93\times$ & 90 & 22 & $4.09\times$ & $7.89\times$ \\ \hline
\end{tabular}
\end{table*}

For Case Study~1, we first train the AC algorithm on three pre-trained scenarios given in the top row of Fig. \ref{fig:SetUp1} and save the converged value functions at each state for each scenario. Then we run both AC and MCAC algorithms on the four deployment scenarios shown in the bottom row of Fig.~\ref{fig:SetUp1}. We use initial state $1$ and goal state $25$ for both pre-trained and deployment scenarios corresponding to Case Study~1. NThe deployment scenarios are obtained by the union of two or more of the obstacles presented in the pre-trained scenarios.

For Case Study~2, we first train the AC algorithm on four pre-trained scenarios given in the top row of Fig. \ref{fig:SetUp2} and save the converged value functions at each state for each scenario. Then we run both AC and MCAC algorithms on deployment scenarios shown in the bottom row of Fig.~\ref{fig:SetUp2}. 
Case Study 2 presents more challenging tasks for the MCAC algorithm 
due to randomly generated obstacles that increase in complexity across the deployment scenarios.

For both environments, we run the AC and MCAC algorithms for 100 episodes, and we conduct 100 independent experiments for each of the pre-trained and deployment scenarios. An episode is defined as the agent starting from the initial state and reaching the goal state.

\subsection{Metrics}
We utilize the following metrics to compare performance of our MCAC algorithm with the AC algorithm. All averages are over 100 independent experiments.

\noindent{\textbf{Average Total Reward:}} The total reward accumulated by the agent in the 100th episode. 

\noindent{\textbf{Average Number of Steps:}} The number of steps taken by the agent to reach the goal in the 100th episode. 

\noindent{\textbf{Average Runtime:}} The amount of time taken to run 100 episodes for each algorithm. 

\noindent{\textbf{Number of Episodes:}} The total number of episodes taken by the AC algorithm to achieve the highest average total reward and the total number of episodes taken by the MCAC algorithm to match or exceed the highest average total reward achieved by the AC algorithm.

\noindent{\textbf{Speed Up - SU1, SU2, and Total SU:}} These metrics quantify efficiency gains of the MCAC algorithm over the AC algorithm. SU1 
is the ratio of the Average Runtime of AC to Average Runtime of MCAC. SU2 
is the ratio of the Number of Episodes of AC to Number of Episodes of MCAC. 
The Total SU is the product of SU1 and SU2, representing the overall speedup achieved by the MCAC algorithm compared to the AC algorithm.

\begin{figure*}[ht]
    \centering
    \begin{subfigure}[b]{0.45\textwidth}
        \includegraphics[width=\textwidth]{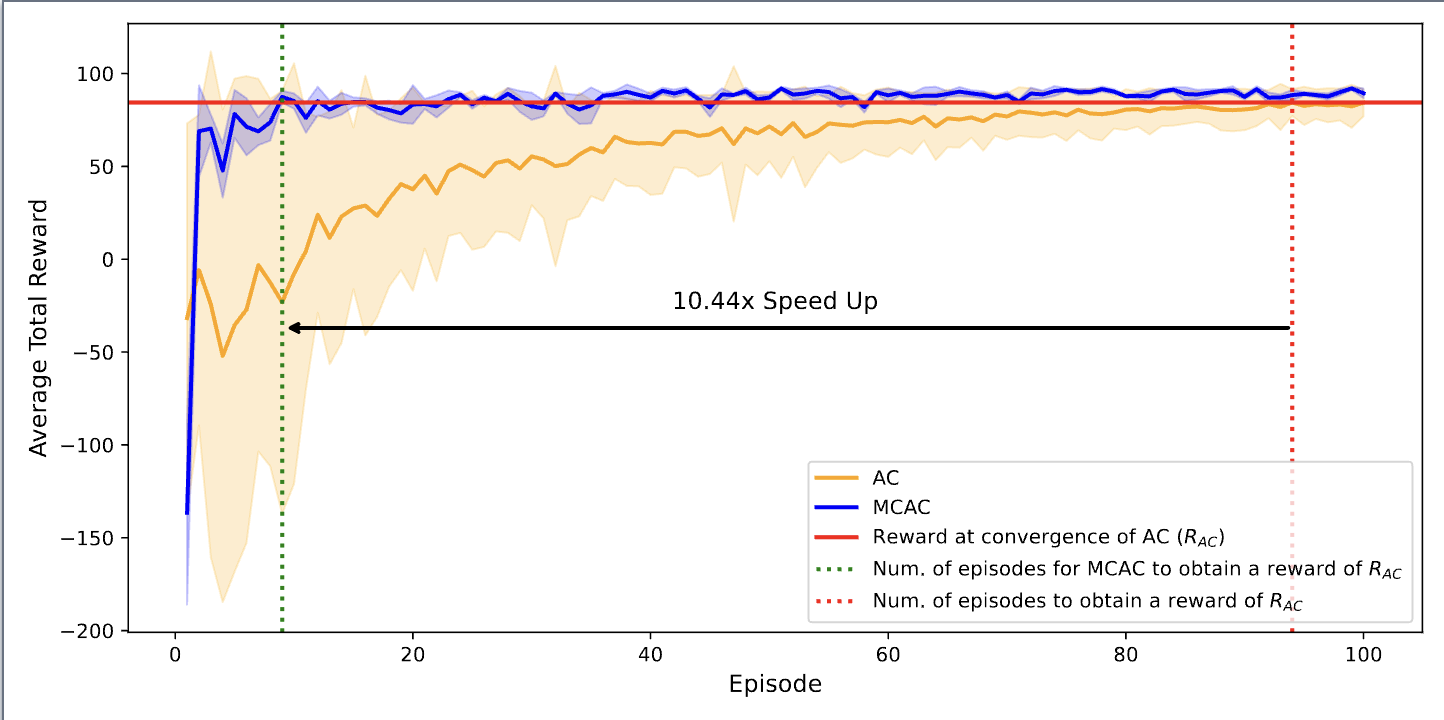}
        \caption{Deployment Scenario 2 - Average Total Reward}
    \end{subfigure}
    \hfill 
    \begin{subfigure}[b]{0.45\textwidth}
        \includegraphics[width=\textwidth]{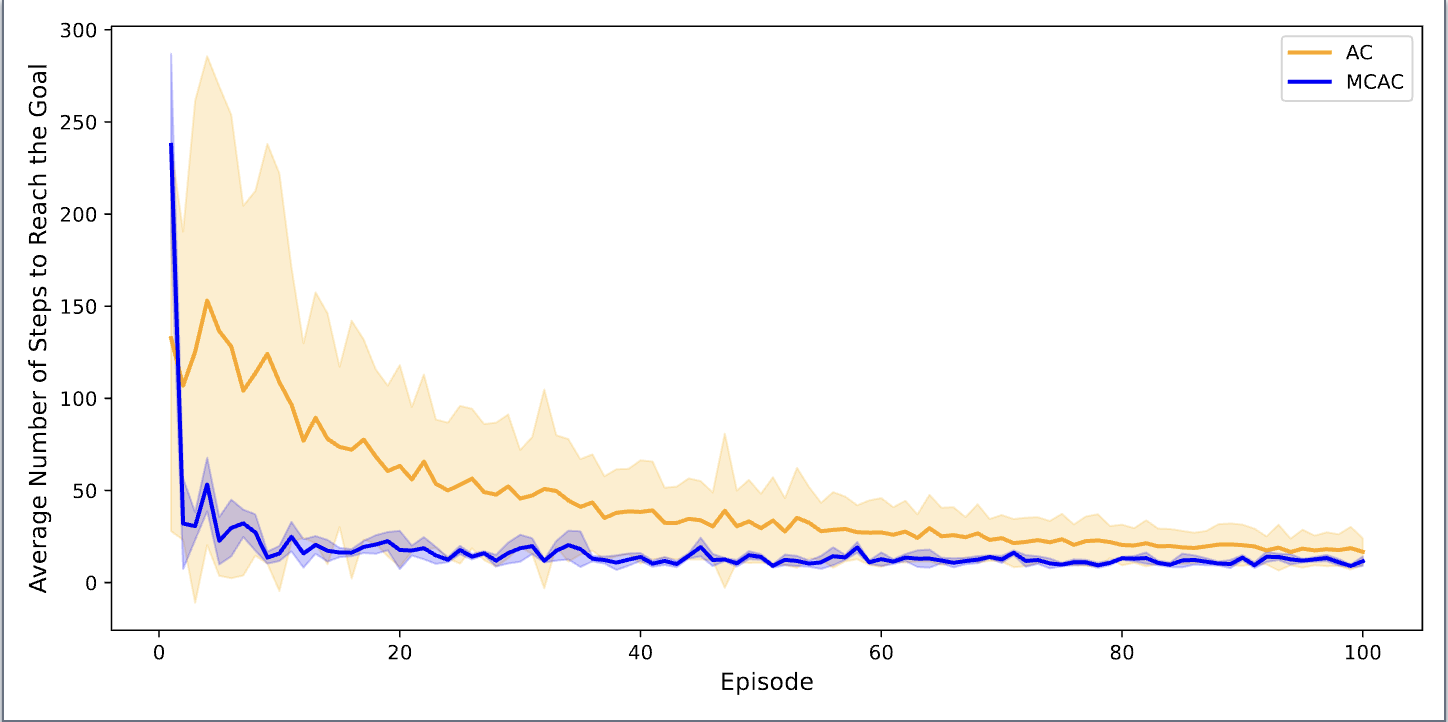}
        \caption{Deployment Scenario 2 - Average Number of Steps}
    \end{subfigure}
    \caption{This figure compares the baseline actor-critic (AC) algorithm and our multi-critic actor critic (MCAC) algorithm in terms of average total reward (left column) and average number of steps to reach the goal (right column) in Deployment Scenario 2 for {\bf Case Study 1}. Shaded regions indicate the variance. MCAC consistently achieves higher average reward and does so in significantly fewer episodes. Using MCAC also results in smaller variance compared to the baseline AC algorithm. Our MCAC algorithm also achieves up to \textbf{10.44x} speedup.
    }\label{fig:ResultsSummary}
\end{figure*}

\subsection{Results}

Table~\ref{table:ResultsCase1} compares performance of our MCAC algorithm against the AC algorithm for Case Study 1. 
The results indicate that the MCAC algorithm consistently achieves a higher average total reward (by $\approx +5$) and enables the agent to reach the goal in fewer steps across all deployment scenarios. 
Our MCAC algorithm also takes less time to convergence (with speedup (SU1) up to \textcolor{black}{2.31x}), and fewer episodes to convergence (with speedup (SU2) up to \textcolor{black}{10.44x}). The total speedup $SU = SU1 \times SU2$ of MCAC is up to \textbf{\textcolor{black}{22.76x}} compared to the baseline AC and \textbf{\textcolor{black}{14.35x}} on average.

Table~\ref{table:ResultsCase2} compares performance of our MCAC algorithm against the AC algorithm for Case Study 1. 
The results indicate that the MCAC algorithm consistently achieves a higher average total reward (by $\approx +140$) and enables the agent to reach the goal in fewer steps across all deployment scenarios. 
Our MCAC algorithm also takes less time to convergence (with speedup (SU1) up to \textcolor{black}{2.06x}), and fewer episodes to convergence (with speedup (SU2) up to \textcolor{black}{4.76x}). The total speedup $SU = SU1 \times SU2$ of MCAC is up to \textbf{\textcolor{black}{9.37x}} compared to the baseline AC and \textbf{\textcolor{black}{8.44x}} on average.

Fig. \ref{fig:ResultsSummary} compares 
the baseline actor-critic (AC) algorithm and our multi-critic actor critic (MCAC) algorithm in terms of average total reward (left column) and average number of steps to reach the goal (right column). 
We observe that MCAC consistently achieves higher average reward and does so in significantly fewer episodes. Using MCAC also results in smaller variance compared to the baseline AC algorithm. 

Our results show that by leveraging pre-existing knowledge of environments through value functions, our MCAC algorithm achieves superior outcomes in terms of optimality and convergence compared to the AC algorithm across 
distinct environments. 
The initial knowledge equips MCAC with an informed starting point, facilitating more effective exploration. Faster convergence of MCAC can be attributed to this improved exploration capability, eliminating a need to train critics from scratch. 
Instead, MCAC fine-tunes weights corresponding to pre-trained critic value functions, efficiently estimating the value of the current deployment environment to identify the optimal solution.

\section{Conclusion}\label{sec:Conclusion}

This paper developed a novel algorithm to facilitate fast autonomy transfer in reinforcement learning. 
Compared to methods that require retraining of parameters when operating in new environments, our multi-critic actor-critic (MCAC) algorithm used knowledge of pretrained value functions and integrated these using a weighted average to expedite adaptation to new environments. 
We demonstrated that MCAC achieved higher rewards and significantly faster (up to \textbf{22.76x}) autonomy transfer compared to a baseline actor-critic algorithm across multiple deployment scenarios in two distinct enrvironments. 
The success of MCAC underscores significance of knowledge transfer in reinforcement learning, offering a promising direction for future research in developing more adaptable and efficient learning systems.
%
\bibliographystyle{IEEEtran}
\bibliography{MultiCritic-Ref}	

\begin{thebibliography}{10}
\providecommand{\url}[1]{#1}
\csname url@samestyle\endcsname
\providecommand{\newblock}{\relax}
\providecommand{\bibinfo}[2]{#2}
\providecommand{\BIBentrySTDinterwordspacing}{\spaceskip=0pt\relax}
\providecommand{\BIBentryALTinterwordstretchfactor}{4}
\providecommand{\BIBentryALTinterwordspacing}{\spaceskip=\fontdimen2\font plus
\BIBentryALTinterwordstretchfactor\fontdimen3\font minus \fontdimen4\font\relax}
\providecommand{\BIBforeignlanguage}[2]{{%
\expandafter\ifx\csname l@#1\endcsname\relax
\typeout{** WARNING: IEEEtran.bst: No hyphenation pattern has been}%
\typeout{** loaded for the language `#1'. Using the pattern for}%
\typeout{** the default language instead.}%
\else
\language=\csname l@#1\endcsname
\fi
#2}}
\providecommand{\BIBdecl}{\relax}
\BIBdecl

\bibitem{sutton2018reinforcement}
R.~S. Sutton and A.~G. Barto, \emph{Reinforcement Learning: {A}n Introduction}.\hskip 1em plus 0.5em minus 0.4em\relax MIT Press, 2018.

\bibitem{hafner2011reinforcement}
R.~Hafner and M.~Riedmiller, ``Reinforcement learning in feedback control,'' \emph{Machine Learning}, vol.~84, pp. 137--169, 2011.

\bibitem{mnih2015human}
V.~Mnih \emph{et~al.}, ``Human-level control through deep reinforcement learning,'' \emph{Nature}, vol. 518, no. 7540, 2015.

\bibitem{silver2016mastering}
D.~Silver \emph{et~al.}, ``Mastering the game of {G}o with deep neural networks and tree search,'' \emph{Nature}, vol. 529, no. 7587, 2016.

\bibitem{zhang2019deep}
C.~Zhang, P.~Patras, and H.~Haddadi, ``Deep learning in mobile and wireless networking: {A} survey,'' \emph{IEEE Communications Surveys \& Tutorials}, vol.~21, no.~3, pp. 2224--2287, 2019.

\bibitem{sadigh2016planning}
D.~Sadigh, S.~Sastry, S.~A. Seshia, and A.~D. Dragan, ``Planning for autonomous cars that leverage effects on human actions.'' in \emph{Robotics: Science and Systems}, 2016.

\bibitem{yan2018data}
Z.~Yan and Y.~Xu, ``Data-driven load frequency control for stochastic power systems: {A} deep reinforcement learning method with continuous action search,'' \emph{IEEE Trans. on Power Systems}, vol.~34, no.~2, 2018.

\bibitem{you2019advanced}
C.~You, J.~Lu, D.~Filev, and P.~Tsiotras, ``Advanced planning for autonomous vehicles using reinforcement learning and deep inverse {RL},'' \emph{Robotics and Autonomous Systems}, vol. 114, pp. 1--18, 2019.

\bibitem{arulkumaran2017deep}
K.~Arulkumaran, M.~P. Deisenroth, M.~Brundage, and A.~A. Bharath, ``Deep reinforcement learning: {A} brief survey,'' \emph{IEEE Signal Processing Magazine}, vol.~34, no.~6, pp. 26--38, 2017.

\bibitem{zhu2023transfer}
Z.~Zhu, K.~Lin, A.~K. Jain, and J.~Zhou, ``Transfer learning in deep reinforcement learning: {A} survey,'' \emph{IEEE Trans. on Pattern Analysis and Machine Intelligence}, 2023.

\bibitem{campos2021beyond}
V.~Campos, P.~Sprechmann, S.~S. Hansen, A.~Barreto, S.~Kapturowski, A.~Vitvitskyi, A.~P. Badia, and C.~Blundell, ``Beyond fine-tuning: Transferring behavior in reinforcement learning,'' in \emph{ICML 2021 Workshop on Unsupervised Reinforcement Learning}, 2021.

\bibitem{watkins1992q}
C.~J. Watkins and P.~Dayan, ``Q-learning,'' \emph{Machine learning}, vol.~8, pp. 279--292, 1992.

\bibitem{konda1999actor}
V.~Konda and J.~Tsitsiklis, ``Actor-critic algorithms,'' \emph{Advances in neural information processing systems}, vol.~12, 1999.

\bibitem{mysore2021multi}
S.~Mysore, G.~Cheng, Y.~Zhao, K.~Saenko, and M.~Wu, ``Multi-critic actor learning: Teaching rl policies to act with style,'' in \emph{International Conference on Learning Representations}, 2021.

\bibitem{li2023multi}
L.~Li, Y.~Li, W.~Wei, Y.~Zhang, and J.~Liang, ``Multi-actor mechanism for actor-critic reinforcement learning,'' \emph{Information Sciences}, vol. 647, p. 119494, 2023.

\bibitem{icarte2022reward}
R.~T. Icarte, T.~Q. Klassen, R.~Valenzano, and S.~A. McIlraith, ``Reward machines: Exploiting reward function structure in rl,'' \emph{Journal of Artificial Intelligence Research}, vol.~73, pp. 173--208, 2022.

\bibitem{icarte2018using}
R.~T. Icarte, T.~Klassen, R.~Valenzano, and S.~McIlraith, ``Using reward machines for high-level task specification and decomposition in reinforcement learning,'' in \emph{International Conference on Machine Learning}.\hskip 1em plus 0.5em minus 0.4em\relax PMLR, 2018, pp. 2107--2116.

\bibitem{icarte2023learning}
R.~T. Icarte, T.~Q. Klassen, R.~Valenzano, M.~P. Castro, E.~Waldie, and S.~A. McIlraith, ``Learning reward machines: A study in partially observable reinforcement learning,'' \emph{Artificial Intelligence}, vol. 323, p. 103989, 2023.

\bibitem{puterman2014markov}
M.~L. Puterman, \emph{Markov decision processes: {D}iscrete stochastic dynamic programming}.\hskip 1em plus 0.5em minus 0.4em\relax John Wiley \& Sons, 2014.

\bibitem{kushner2012stochastic}
H.~J. Kushner and D.~S. Clark, \emph{Stochastic approximation methods for constrained and unconstrained systems}.\hskip 1em plus 0.5em minus 0.4em\relax Springer Science \& Business Media, 2012, vol.~26.

\bibitem{metivier1984applications}
M.~Metivier and P.~Priouret, ``Applications of a {K}ushner and {C}lark lemma to general classes of stochastic algorithms,'' \emph{IEEE Trans. on Information Theory}, vol.~30, no.~2, pp. 140--151, 1984.

\bibitem{dong2020survey}
X.~Dong, Z.~Yu, W.~Cao, Y.~Shi, and Q.~Ma, ``A survey on ensemble learning,'' \emph{Frontiers of Computer Science}, vol.~14, pp. 241--258, 2020.

\bibitem{wortsman2022model}
M.~Wortsman, G.~Ilharco, S.~Y. Gadre, R.~Roelofs, R.~Gontijo-Lopes, A.~S. Morcos, H.~Namkoong, A.~Farhadi, Y.~Carmon, S.~Kornblith \emph{et~al.}, ``Model soups: {A}veraging weights of multiple fine-tuned models improves accuracy without increasing inference time,'' in \emph{International Conference on Machine Learning}.\hskip 1em plus 0.5em minus 0.4em\relax PMLR, 2022, pp. 23\,965--23\,998.

\bibitem{borkar2009stochastic}
V.~S. Borkar, \emph{Stochastic {A}pproximation: {A} {D}ynamical {S}ystems {V}iewpoint}.\hskip 1em plus 0.5em minus 0.4em\relax Springer, 2009, vol.~48.

\end{thebibliography}
	
\end{document}